\documentclass{article} 
\usepackage{iclr2023_conference,times}

\usepackage[utf8]{inputenc} 
\usepackage[T1]{fontenc}    
\usepackage{hyperref}       
\usepackage{url}            
\usepackage{booktabs}       
\usepackage{amsfonts}       
\usepackage{amsthm}
\usepackage{nicefrac}       
\usepackage{microtype}      
\usepackage{xcolor}         

\usepackage[pdftex]{graphicx}
\usepackage{caption}
\usepackage{subcaption}
\usepackage{hyperref}
\hypersetup{
    colorlinks=true,
    citecolor=black,
    linkcolor=black,
    urlcolor=blue,
    }
\usepackage{algorithm}
\usepackage{algpseudocode}
\usepackage{amsmath}

\usepackage{xspace}

\usepackage{textgreek}
\usepackage[normalem]{ulem}

\mathchardef\mhyphen="2D

\title{Multipath agents \\ for modular multitask ML systems}

\author{
    Andrea Gesmundo \\
    Google Research \\
    \texttt{agesmundo@google.com} \\
}

\begin{document}

\maketitle

\begin{abstract}
A standard ML model is commonly generated by a single method that specifies aspects such as architecture, initialization, training data and hyperparameters configuration.
The presented work introduces a novel methodology allowing to define multiple methods as distinct agents.
Agents can collaborate and compete to generate and improve ML models for a given tasks.
The proposed methodology is demonstrated with the generation and extension of a \emph{dynamic modular multitask ML system} solving more than one hundred image classification tasks.
Diverse agents can compete to produce the best performing model for a task by reusing the modules introduced to the system by competing agents.
The presented work focuses on the study of agents capable of: 1) reusing the modules generated by concurrent agents, 2) activating in parallel multiple modules in a frozen state by connecting them with trainable modules, 3) condition the activation mixture on each data sample by using a trainable router module.
We demonstrate that this simple per-sample parallel routing method can boost the quality of the combined solutions by training a fraction of the activated parameters.

\end{abstract}

\section{Introduction}
This work extends the $\mu$Net line of research \citep{Gesmundo2022munet1,Gesmundo2022munet2,Gesmundo2022munet3,Gesmundo2022munet4}.
This line of research aims to define and demonstrate novel methodologies to enable the creation of dynamic large-scale multi-modal multitask modular intelligent systems that can be indefinitely and collaboratively extended.
Such intelligent systems can accelerate and automate Machine Learning (ML) development and innovation by enabling higher quality solutions with higher efficiency/automation and lower the entry barrier for less specialized users.

This work demonstrates the \emph{method heterogeneity} and \emph{collaborative ML development} capabilities enabled by the \emph{multiagent framework} introduce by \citet{Gesmundo2022munet4}. 
The use of collaborative agents 
is demonstrated by applying the \emph{continual development methodology} \citep{Gesmundo2022munet3} to extend the $\mu$3Net system \citep{Gesmundo2022munet4}.
The $\mu$3Net system is a multitask system solving jointly 124 image classification tasks.
This system is composed by a set of modules/components and a set of models/paths.
Different paths can share modules.
Each path connects a subset of modules to define a model for a task.
Such models/paths are generated by multiple installations of a \emph{singlepath agent} (i.e. one instantiation per task).
This publication describes the extension of such system into a ``Mutant Multiagent Multipath Multitask Network'' ($\mu$4Net).
$\mu$4Net introduces a novel \emph{multipath agent} capable of generating architectures activating multiple parallel paths.
The multipath agent can be instantiated by specifying an assigned task.
Each multipath agent combines available models/paths in a frozen state (i.e. with no additional training). The frozen paths are connected with trainable connector modules.
Furthermore, the aggregation of the representations generated by the parallel paths is controlled by a router module. The router learns to weigh the activated paths conditioning on features of each input sample.

The following sections define the method implemented by the multipath agent, and report an empirical study aimed to analyze the properties and motivate the novel design elements.

\section{Method}
\label{section:method}

This section details the method implemented by the proposed multipath agent.

\subsection{Multipath architecture}
\label{sec:arch}

\begin{figure}[t]
  \centering
  \includegraphics[width=0.9\linewidth]{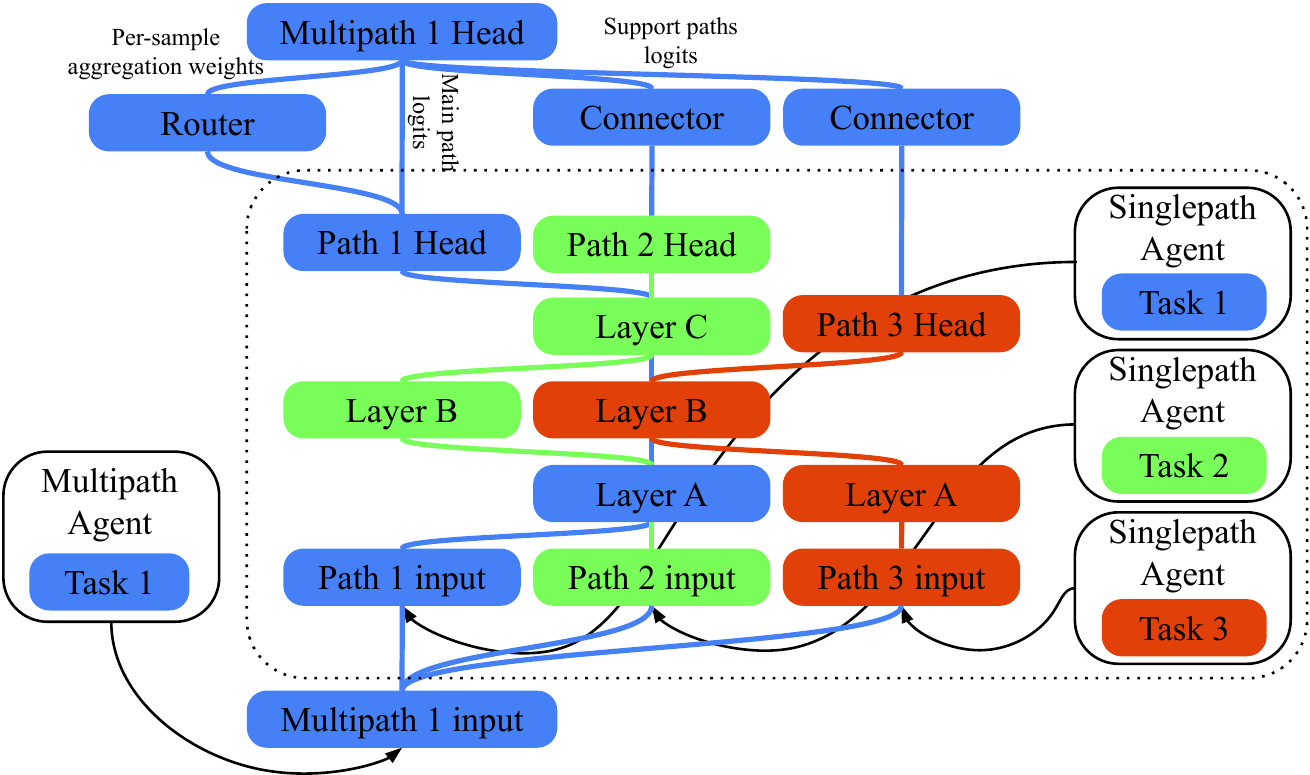}
  \caption{
  Representation of an exemplar architecture generated by a \emph{multipath agent} within a \emph{modular multitask system} solving 3 tasks. Each task is assigned to an instance of a singlepath agent. Each singlepath agent generates a path/model solving the assigned task. Paths can share modules/knowledge if beneficial. An additional multipath agent generates an improved solution for Task 1 by augmenting the logits produced by the path solving Task 1 (\emph{main path}). The main logits are augmented with the representations produced by support paths. The support representations are mapped into the target logits space by using connector modules. The aggregation of the logits is performed by a per-sample router conditioned on the main path logits. The multipath agent cannot alter the activated paths and can train only the router and connectors modules.
  }
\label{fig:arc}
\end{figure}

The proposed multipath agent generates architectures designed to provide an effective way to boost the quality achieved by the model/path generated by the singlepath agent solving the same task.
Figure~\ref{fig:arc} represents the type of multipath architectures generated by the multipath agent.
These multipath architectures are composed by:
\begin{enumerate}
    \item \emph{Main path}. The main path is the model generated by the singlepath agent solving the same task. The main path is used in a frozen state (i.e. cannot be altered during the evolution or training of the multipath structures).
    \item \emph{Support paths}. An arbitrary number of additional paths can be activated in parallel to improve the main path. Support paths are also used in a frozen state. In general, the only requirement for support paths is to be capable of processing the input samples provided by the target task. In the application considered (i.e. extension of the $\mu$3Net system), all the existing paths solve image classification tasks. Thus any path in the system can be used as a support path with minimal pre-processing adaptations (i.e. image resizing).
    \item \emph{Connector modules}. The representation produced by each support path provides additional information to improve the main model/path.
    In general, this additional information can be used to augment any internal representation of the main model. In the instantiation presented, the support representations are used to improve the logits of the main path ($\mu$3Net solves jointly 124 image classification tasks each with different label set).
    Trainable connectors modules are used to map the representation produced by each support path into the logits space of the target task. Therefore, connectors are required to change the shape of the representation tensor and map its semantic. In the proposed instantiation, the connector modules are implemented as \textbf{a single fully connected layer} (FCL). The connectors are zero initialized. Therefore, the output of the multipath models is initially determined only by the main path, as the initialized connectors output zeros.
    \item \emph{Router module}. The logits produced by the main path are aggregated with the logits produced by the connected support paths. The aggregation is performed by a weighted average of the different logits tensors. The weights are provided by a trainable router module. The weights are sample specific. Per-sample weighting can grant more influence to paths providing specialized knowledge relevant to each example.
    The router needs to be conditioned on the input sample to be able to produce per-sample weights. In general, any representation of the input sample can be used. In the proposed instantiation, the router is conditioned on the main path logits.
    The router is also implemented as \textbf{a single fully connected layer}.
    A softmax is applied to the output of the FCL to produce normalized weights.
    All the parameters of the router's FCL are zero initialized except for the bias value corresponding to the main path weight, that is set to impose a prior over the main path weight, as detailed in Section~\ref{sec:init_bias}.
    Furthermore, the routing weights are applied only during the forward pass, while gradients are computed as if the logits are aggregated with a sum as detailed and motivated in Section~\ref{sec:bdr}
\end{enumerate}

Many multipath agents can be executed in parallel and asynchronously, each assigned to a distinct target task. 
The use of a path can be shared across many multipath architectures as the paths are always used in a frozen state.
Notice that the implementation of such multipath architecture can be rendered efficient by caching the representations produced by the paths (i.e. each input is always mapped to the same representation as single paths are frozen).
Furthermore the execution of a support path can be skipped if the router produces a weight close to zero for its representation (i.e. in this case, the representation can be replaced with zero padding, as it would not be a relevant factor of the aggregation).

\begin{figure}[t]
  \vspace{-15pt}
  \centering
  \includegraphics[width=1.\linewidth]{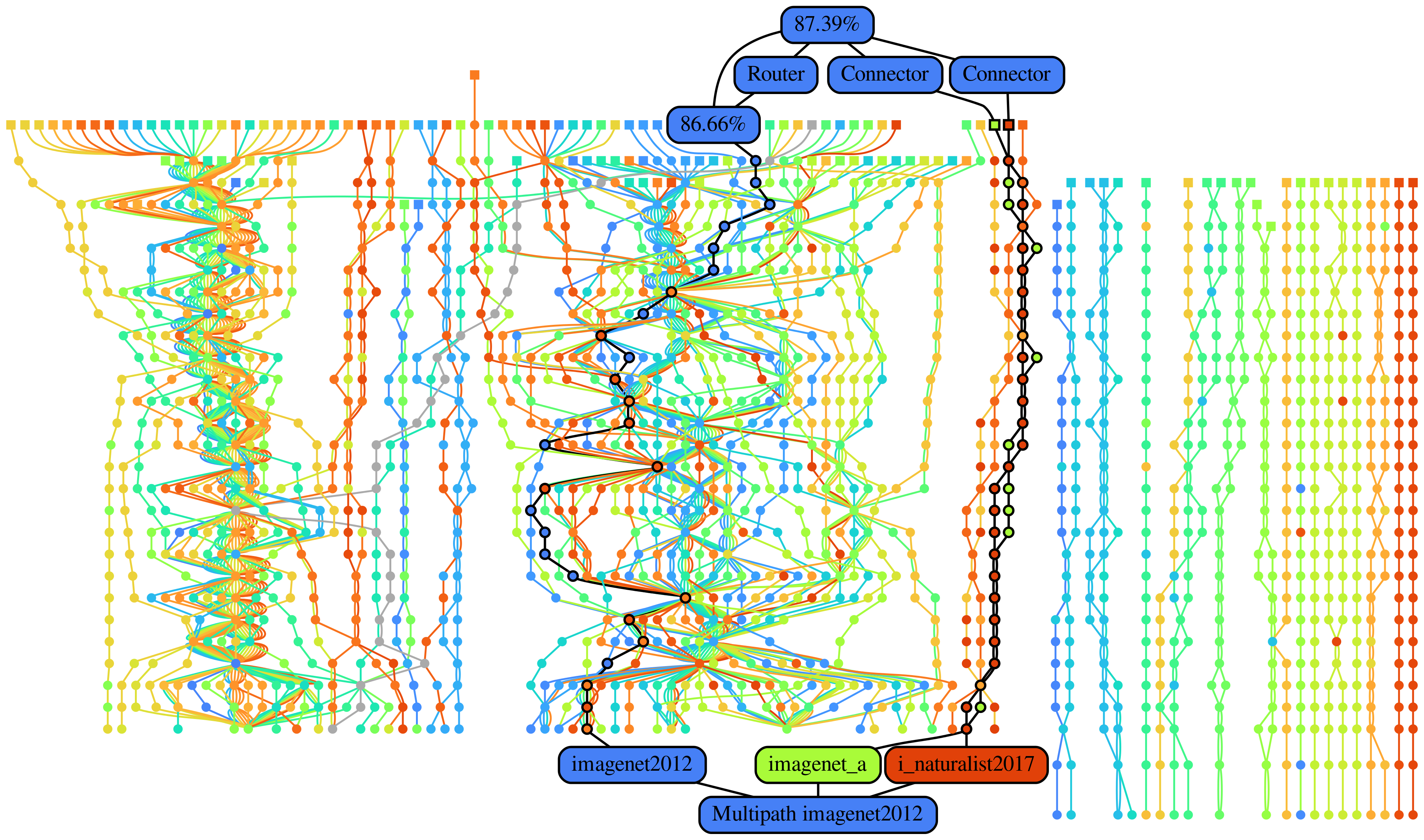}
  \vspace{-15pt}
  \caption{
Graph representing $\mu$3Net \citep{Gesmundo2022munet4} with an additional multipath structure (highlighted with black edges) providing an improved solution for the 
\href{https://www.tensorflow.org/datasets/catalog/imagenet2012}{imagenet2012} task.
The single path model achieve 86.66\% test accuracy while the multipath model achieves 87.39\% by training only 3 fully connected layers (i.e. 1 for the router and 2 for the connector modules).
This multipath model leverages the representation generated by the paths trained on the tasks \href{https://www.tensorflow.org/datasets/catalog/i_naturalist2017}{i\_naturalist2017} and \href{https://www.tensorflow.org/datasets/catalog/imagenet_a}{imagenet\_a}.
Each task is identified with a unique color.
Top rectangular nodes represent the head module of each task.
Each sequence of edges of the same color defines the sequence of modules composing each model.
Internal nodes are represented with the color of the task on which the parameters of the corresponding module were fine-tuned last.
Input nodes of highlighted paths report the correspondent task name.
Input nodes of non-highlighted paths are omitted. 
}
\label{fig:arc-large}
\end{figure}

\subsubsection{Router bias initialization}
\label{sec:init_bias}

This section describes the technique used to initialize the parameters of the router's FCL in order to achieve the following properties:
\begin{enumerate}
    \item Set the initial routing weight assigned to the main path to: ${w_{main}^* \in [0, 1]}$.
    \item Distributes the remaining weight uniformly across the support paths: ${w_{support_n} = (1-w_{main}^*)/{(|\mathcal{P}|-1)}}$, where $\mathcal{P}$ is the set of activated paths (i.e. main path $\cup$ support paths) and ${n \in [1, |\mathcal{P}\!-\!\{main\}|]}$.
\end{enumerate}
The initialization method proposed in this section guarantees these properties to be valid for the initialized state of any multipath architecture, regardless of the number of activated support paths.

These properties are achieved by zero initializing all the parameters of the router's FCL with exception of the bias value corresponding to the main path, which needs to be set to:
\begin{equation} \label{eq:1}
b_{main}^{init} = \ln\left(\frac{1}{(\frac{1}{w_{main}^*}-1)}\cdot(|\mathcal{P}|-1)\right)
\end{equation}

\begin{proof} 
The computation producing the routing weights can be expressed as:
$$ \mathbf{w} := \left| \begin{array}{c} w_{main} \\ w_{support_1} \\ \vdots \\ w_{support_{|\mathcal{P}|-1}} \end{array}  \right| = softmax(W \cdot \mathbf{x} + \mathbf{b})$$
Where $\mathbf{x}$ is the tensor inputted to the router's FCL, $\mathbf{b}$ is the bias vector of the router's FCL, and $W$ is the parameters matrix of its kernel.
Considering that $W$ is zero initialized, the initial state of the routing weights is simplified to:
$ \mathbf{w}^{init} = softmax(\mathbf{b}^{init})$.
By applying the softmax formula, the initial value of $w_{main}$ results:
\begin{equation}
w_{main}^{init} = \frac{e^{b_{main}^{init}}}{e^{b_{main}^{init}} + \sum_{n=1}^{|\mathcal{P}|-1} e^{b_{support_n}^{init}}}
\end{equation}
Considering that all the bias values corresponding to support paths are zero initialized (i.e.~${b_{support_n}^{init}\!= 0 \ \ \forall\ n \in [1, |\mathcal{P}\!-\!\{main\}|]}$):
\begin{equation}w_{main}^{init} = \frac{e^{b_{main}^{init}}}{e^{b_{main}^{init}} + |\mathcal{P}|-1}
\end{equation}
Finally, $b_{main}^{init}$ can be substituted by applying equation~(\ref{eq:1}):
\begin{equation}
w_{main}^{init}\!=\!\frac{\left(\frac{1}{(\frac{1}{w_{main}^*}-1)}\cdot(|\mathcal{P}|\!-\!1)\right)}{\left(\frac{1}{(\frac{1}{w_{main}^*}-1)}\cdot(|\mathcal{P}|\!-\!1)\right) + |\mathcal{P}|\!-\!1}\!=\!\frac{\frac{1}{(\frac{1}{w_{main}^*}-1)}}{\frac{1}{(\frac{1}{w_{main}^*}-1)} + 1}\!=\!\frac{1}{1+(\frac{1}{w_{main}^*}-1)}\!=\!w_{main}^*
\end{equation}

Also the initial routing weight of support paths matches with the stated properties, $\forall n \in [1, |\mathcal{P}|-1]$:
\begin{equation}w_{support_n}^{init}\!\!=\!\frac{e^{b_{support_n}^{init}}}{e^{b_{main}^{init}}\!+\!\sum_{n=1}^{|\mathcal{P}|-1}e^{b_{support_n}^{init}}}\!=\!\frac{\frac{1}{|\mathcal{P}|-1}}{\frac{1}{(\frac{1}{w_{main}^*}-1)}\!+\!1}\!=\!\frac{\frac{1}{|\mathcal{P}|-1}}{\frac{\frac{1}{w_{main}^*}}{\frac{1}{w_{main}^*}-1}}\!=\!\frac{\frac{1}{|\mathcal{P}|-1}}{\frac{1}{1-w_{main}^*}}\!=\!\frac{1\!-\!w_{main}^*}{|\mathcal{P}|-1}
\vspace{8pt}
\end{equation}
\end{proof}
\clearpage

During preliminary experiments, we have observed that initializing with a higher $w_{main}^*$ can have two visible effects:
\begin{enumerate}

\item The quality of an initialized multipath model matches of the quality achieved by the frozen main path (also considering that connectors are zero initialized). The higher $w_{main}^*$ is set
the less likely it is that the quality of the multipath model drops below the main path quality during training.
Conversely, a too high $w_{main}^*$ can slow the learning of the routing patterns required to leverage the additional information provided by the support paths.
\item Faster converge of the multipath model training.
This finding matches with the main design intuition as the main path is expected to be the most influential in average.
In fact, the value used in the following experiments is set to a value within the range that have been measured at convergence of preliminary experiments: ${w_{main}^*\!=80\%}$.
\end{enumerate}

\subsubsection{Backprop decoupled routing}
\label{sec:bdr}

In general, routing methods rely on the aggregation of alternative representations produced by different modules or experts. A common application of this design pattern are mixture of experts architectures \citep{Shazeer2017OutrageouslyLN}.
The routers are usually trained jointly with the modules whose output is aggregated.
During the forward pass, the routing weights are applied to select the most relevant representation/module.
The weighting has also the effect of scaling the gradients that modules received during backward propagation (i.e. assigning a low weight to a module will results in proportionally diminished gradients).
Thus, routers tend to produce distributions that are excessively skewed toward a few modules that perform better during early training. In practice, the training of unselected modules grinds to a halt because their gradients vanishes as the ``quality'' gap with respect to the trained modules increases.
Thus, routing methods can often suffer from load balance issues and suboptimal quality.
This common issue has be referred to as the \emph{rich~gets~richer} effect \citep{Shen2019MixtureMF}.
This effect is more accentuated for the proposed multipath method, because the main path is initialized with a converged model, while the support paths require training of the connector.
Common solutions rely on auxiliary loss factors such as \emph{entropy regularization} of the routing weights distribution to reduce its skewness.
Such solutions add complexity as additional logic is required to gather and propagate through the network the information needed for the auxiliary loss factors computation. Furthermore, auxiliary loss factors require to tune additional hyperparameters such as the scaling parameters.

In this section we introduce an alternative solution that does not require additional hyperparameters and can be defined and implemented within the weighted aggregation logic.
The proposed technique is based on applying different aggregations for the forward pass and backward pass.

The standard router weighted aggregation can be represented as:
\begin{equation} \label{eq:2}
\mathbf{a} = R \cdot \mathbf{w} = \left| 
\begin{array}{c} \mathbf{r}_{1} \\ \vdots \\ \mathbf{r}_M \end{array} 
\right| \cdot \mathbf{w}
\end{equation}
Where $\mathbf{w}$ is the vector of routing weights, $M$ is the number of modules whose representations are being aggregated, $\mathbf{r}_i$ denotes each representation, and $R$ represents the matrix obtained by stacking the representations.
The proposed technique consists in applying the standard aggregation during the forward propagation but applying an unweighted aggregation during the backward propagation:
\begin{equation} \label{eq:3}
\mathbf{a}_{backprop} = \left| 
\begin{array}{c} \mathbf{r}_{1} \\ \vdots \\ \mathbf{r}_M \end{array} 
\right| \cdot \mathbf{1}
\end{equation}
Notice that this allows the modules to receive unscaled gradients, independently from the selection choice of the router.
This property is particularly critical for the instantiation of the multipath method proposed, as the connectors can be trained even if the router naturally tends to converge to select the main path during early training.

Let $stopgradient(\cdot)$ represent a function that acts as the identity function during the forward propagation but stops the flow of gradients during the backward propagation:
\begin{equation}
stopgradient(x) = \begin{cases}
x,\ \ \ \text{if forward propagation}\\
0,\ \ \ \text{if backward propagation}
\end{cases}
\end{equation}
The forward and backward propagation formulas (\ref{eq:2} and \ref{eq:3}) can be represented jointly by using the ${stopgradient(\cdot)}$ function:
\begin{equation}
\label{eq:bdr}
 \mathbf{a}_{joint} = stopgradient(R)\cdot\mathbf{w} + R\cdot\mathbf{1} - stopgradient(R\cdot\mathbf{1})
\end{equation}
During the forward propagation: ${\mathbf{a}_{joint} = R\cdot\mathbf{w}}$, since $stopgradient(\cdot)$ has no effect in this phase and the remaining two factors cancel out: ${(R\cdot\mathbf{1}-stopgradient(R\cdot\mathbf{1})) = 0}$.
While, during the backward propagation, the modules receive gradients only through the factor ${R\cdot\mathbf{1}}$, that is unaffected by the routing weights.
Notice that the router is still trained following the standard aggregation formula~(\ref{eq:2}) as the router receives gradients only through the factor: ${stopgradient(R)\cdot\mathbf{w}}$.

Notice that, the used formulation does not allow the modules to specialize on the subset of samples for which the router selects them with higher weight.
For applications where specialization may be more important, it is possible to define alternative solutions based on the same backprop decoupling intuition.
For example: 
\begin{equation}
\mathbf{a}'_{backprop} = \left| 
\begin{array}{c} \mathbf{r}_{1} \\ \vdots \\ \mathbf{r}_M \end{array}
\right| \cdot 
\left| \begin{array}{c} w_{1}/EM\!A(w_1) \\ \vdots \\ w_N/EM\!A(w_N) \end{array}\right|
\end{equation}
Where each weight is scaled by its exponential moving average: $EM\!A(\cdot)$. Notice that this solution still avoids the \emph{rich gets richer} effect, since $\mathbb{E} [w_i/EM\!A(w_i)] \cong 1 $ even if $\mathbb{E}[w_i]\rightarrow 0$.

\subsubsection{Router learning rate multiplier}
\label{sec:lrm}
The learning rate applied to the router module is allowed to be set independently from the global learning rate applied to the other components.
This is parametrized by adding a new hyperparameter: \emph{router learning rate multiplier}.
This hyperparameter determines the learning rate applied to the router by scaling the global learning rate:
\begin{equation}
\label{eq:lrs}
learning\mbox{-}rate_{router} =  \lambda \cdot learning\mbox{-}rate
\end{equation}
Where $\lambda$ represents the router learning rate multiplier hyperparamter.
This multiplicative formalization allows for a learning rate scaling implementation that is self-contained within the router module by using the $stopgradient(\cdot)$ function:
\begin{equation}
\label{eq:lrs2}
    \mathbf{o}_{grad\mbox{-}scaled} = \lambda \cdot \mathbf{o} + (1 - \lambda) \cdot stopgradient(\mathbf{o})
\end{equation}
Where $\mathbf{o}$ represents the output tensor of the router module.
During the forward pass: ${\mathbf{o}_{grad\mbox{-}scaled} = \mathbf{o} \ \ \ \forall \ \ \lambda \in \mathbb{R}}$.
While, during the backward propagation, the gradients applied are scaled by $\lambda$ as: ${\mathbf{o}_{grad\mbox{-}scaled} = \lambda \cdot \mathbf{o}}$.

The incremental auto-tuning method \citep{Gesmundo2022munet1} can be applied also to $\lambda$.
The need for a distinct learning rate for the router and its default value and auto-tuning range have been identified during preliminary experiments (see Table~\ref{table:hps}).

Notice that, the presented \textbf{modular learning rate scaling} technique is generally applicable to any module, as the logic expressed by equation~(\ref{eq:lrs2}) can be applied to the output of any module. For example it can be applied to modules of the models generated by the singlepath agent to scale the learning rate of each module independently.

\subsection{Multipath models sampling}
\label{sec:mms}
In this section we define the method used to sample multipath models.
This sampling method is intentionally designed to be as similar as possible to the singlepath agent's model sampling \citep{Gesmundo2022munet4}, in order to demonstrate its generality and allow for a fair comparison.
In this section we give a summary of the method, for further details refer to prior work and published code.

\subsubsection{Parent model selection}
New multipath models are sampled by mutation. We refer to the model being mutated as the \emph{parent model}. The agent attempts to sample the parent model from a set of  multipath models previously sampled and scored by the same agent. We refer to this set of candidate parent models as the \emph{population}. 
The models in the population are visited in decreasing order of score, starting with the highest scoring one.
Each model, $m$, can be accepted as parent at random with probability:
\begin{equation}
p_{parent}(m) = 0.5^{\#of\!fsprings(m)}
\end{equation}
Where $\#of\!fsprings(m)$ denotes the number models that have been previously generated by using $m$ as parent.
If the current candidate parent is not selected, then iteratively the model with the next best score is considered to be selected as parent with probability $p_{parent}(\cdot)$.
If a full iteration on the population is completed without a successful parent model selection (i.e. the population is empty or all candidates are rejected), then a \emph{randomly initialized multipath model} is used as parent.
This random model is sampled as follow:
\begin{enumerate}
    \item its hyperparameters are copied from the configuration of the highest scoring model in the population.
    If the population is empty, then the hyperparameters are set to the default values (see Table~\ref{table:hps}).
    \item its support paths are uniformly sampled without replacement from the set of the single paths available in the system.
    \item the connectors are zero initialized and the router is initialized with the bias initialization technique defined in Section~\ref{sec:init_bias}.
\end{enumerate}


\begin{table}[t]
  \caption{
Hyperparameters search space.
Sequences of valid values for each automatically tunable hyperparameter.
Bold values are defaults.
Underlined values highlight the differences with the search space used by the single path agent \citep{Gesmundo2022munet3}.
The image resolution cannot be tuned by the multipath agent, since each activated path is used in a frozen state and needs to be provided with an image resolution matching its configuration.
An additional hyperparameter has been added to scale the learning rate applied to the router module.}
  \label{table:hps}
  \centering
  \begin{tabular}{l}
\toprule
\multicolumn{1}{c}{\emph{Optimizer hyperparameters}} \\
Learning rate $\in$ [0.0001, 0.0002, 0.0005, 0.001, 0.002, 0.005, 0.01, \textbf{0.02}, 0.05, 0.1, 0.2, 0.5] \\
Learning rate schedule warm up ratio $\in$ [0, 0.01, \textbf{0.02}, 0.05, 0.1, 0.2, 0.3] \\
Momentum $\in$ [0.5, 0.6, 0.7, 0.75, \textbf{0.8}, 0.85, 0.9, 0.95, 0.98, 0.99] \\
Nesterov update $\in$ [False, \textbf{True}] \\
\underline{Router learning rate multiplier $\in$ [0.01, 0.02, \textbf{0.05}, 0.1, 0.2, 0.5, 1]}
\\
\midrule
\multicolumn{1}{c}{\emph{Data Preprocessing hyperparameters}} \\
Cropped area range min $\in$ [0.05, 0.5, 0.95, \textbf{1.0}] \\
Cropped aspect ratio range min $\in$ [0.5, 0.75, \textbf{1.0}] \\
Flip left/right $\in$ [\textbf{False}, True] \\
Brightness delta $\in$ [\textbf{0.0}, 0.01, 0.02, 0.05, 0.1, 0.2] \\
Contrast delta $\in$ [\textbf{0.0}, 0.01, 0.02, 0.05, 0.1, 0.2] \\
Saturation delta $\in$ [\textbf{0.0}, 0.01, 0.02, 0.05, 0.1, 0.2] \\
Hue delta $\in$ [\textbf{0.0}, 0.01, 0.02, 0.05, 0.1, 0.2] \\
Image quality delta $\in$ [\textbf{0.0}, 0.01, 0.02, 0.05, 0.1, 0.2] \\
\underline{\sout{Image resolution pixels $\in$ [224, \textbf{384}]}} \\
\bottomrule
  \end{tabular}
\vspace{-12pt}
\end{table}
\subsubsection{Model mutation}

Two types of model mutations are available:
\begin{enumerate}
    \item \emph{Hyperparameter mutation}. Hyperparameters of the optimizer and preprocessing can be mutated. The hyperparameter set and their value defaults and ranges is equivalent to the one used by the singlepath agent \citep{Gesmundo2022munet3} with 2 exceptions: 1) image resolution tuning is disabled since each activated path is used in a frozen state and needs to be provided with an image with resolution matching its training configuration.
    2) an additional optimizer hyperparameter has been added to scale the learning rate applied to the router module.
    The hyperparameters value changes are sampled by using the same \emph{incremental hyperparameter mutation} technique used by the singlepath agent \citep{Gesmundo2022munet1}.
    The default value and auto-tuning range of the \emph{router learning rate multiplier} have been identified during preliminary experiments. 
    The default values for each of the remaining hyperparameters is set to the most frequently used value by the models included in the $\mu$3Net system (see \citet{Gesmundo2022munet4} Figure~6).
    \item \emph{Support path mutation}. The set of support paths used by the parent model can be altered by adding or removing one support path. If a support path is added, then a new single path is sampled (avoiding duplicates) and a new zero-initialized connector module is added. If a support path is removed, then its connector is removed and the number of activated support paths is reduced by one. In both cases the parameters of the router are reset.
    Any trainable component inherited from the parent model is cloned in order to guarantee immutability of the pre-existing models (equivalent to the layer cloning mutation used by singlepath agents \citet{Gesmundo2022munet3}). For the experiments described in this paper, the default number of used paths is 2 and the maximum is 3. 
\end{enumerate}

The selected parent model can be mapped to a newly sampled multipath model by applying a subset of the possible mutation actions.
Any candidate mutation is selected for application with probability $\mu(\delta|m)$, where $\delta$ represents the candidate mutation, $m$ is the parent model, and $\mu(\cdot)$ is the learned $\mu$ function as defined in \citet{Gesmundo2022munet3}.


\subsubsection{Evolutionary cycles}
The logic of the evolutionary cycles and the specific configuration used in the following experiments matches the logic used by the singlepath agent \citep{Gesmundo2022munet4}.

Each multipath agent searches for the best scoring architecture by performing evolutionary cycles.
At the end of each cycle only the best scoring model is retained in the system and evaluated on the test set, while other models are discarded.
During each cycle, 16 models are sampled with the method described in Section~\ref{sec:mms}.
Each agent samples and trains 4 models in parallel.
Each model is trained on one of the 4 TPUv4 MegaCore chips available.
During training, the model is evaluated on the validation set 4 times at regular intervals.
The quality measured on the validation set is used as score.
Only the checkpoint corresponding to the intermediate evaluation achieving the best score is retained.
The scored models are then inserted in the agent's population, $\mathcal{P}$, to be used as candidate parent and possibly become the new best scoring model.
Early population pruning is performed by discarding the models that did not achieve a better score than their parent \citep{Gesmundo2022munet1}.

\begin{table*}[t]
\begin{center}
\caption{
Comparison between the quality achieved by the proposed multipath agent and alternative methods on the \href{https://www.tensorflow.org/datasets/catalog/imagenet2012}{imagenet2012} task.
Both the validation accuracy (used as score by the $\mu$Net agents) and the test accuracy (computed on a held-out dataset) are reported.
Each multipath agent experiment is repeated 10 times and lasts 15 evolutionary cycles.
The mean of the 10 accuracies and its standard error (s.e.m.) is reported for each multiagent experiment. The maximum of the 10 test accuracies is also reported.
As baselines, the mean test accuracy achieved by \citet{Dosovitskiy2021AnII} with 3 fine-tuning repetitions of a ViT Large model matching pretraining and size of the $\mu$Net root model is reported.
For the singlepath agent baseline, the accuracy achieved by the path included in the $\mu$3Net version of the system is reported in the ``mean'' column, while the best test accuracy recorded through the whole history of the $\mu$Net system is reported in the ``max'' column.
The \emph{Ablation study} and \emph{Headroom analysis} sections report results achieved by modifications of the multipath agent method as described in Sections~\ref{ssec:ablation} and \ref{ssec:headroom}.
For each set of ablation and headroom experiments, the significance of the test accuracy deltas achieved with respect to the unmodified multipath agent is measured as \emph{p}-value computed with the \emph{independent two-sample t-test}.
 with respect to the multipath agent experiment repetitions.
Plots representing the change over time of these metrics are reported in Figure~\ref{fig:curves}.
Results are discussed in Section~\ref{section:experiments}.
}
\label{table:bestmetrics}
\begin{tabular}{lcccc}
\toprule
 Method & \multicolumn{1}{c}{Validation acc. {\small(\%)}} & \multicolumn{3}{c}{Test accuracy {\small(\%)}}\\
                    & mean\ {\small$\pm$s.e.m.} 
                    & mean\ {\small$\pm$s.e.m.} 
                    & max 
                    & \emph{p}-value
                    \\
\midrule
    Multipath agent 
    &  79.25\ {\small$\pm$0.05} 
    & 87.19\ {\small$\pm$0.03}
    & 87.29
    \\
\midrule
\multicolumn{5}{c}{\emph{Baseline methods}} \\
    ViT Large fine-tuning
        & 
        & 85.30\ {\small$\pm$0.01}
        \\
    Singlepath agent 
    &  78.54 \hspace{23.4pt} 
    & 86.66 \hspace{23.4pt} 
    & 86.74
    \\
\midrule
\multicolumn{5}{c}{\emph{Multipath agent: Ablation study}} \\
    Ablate backprop decoupled routing
                    &  78.65\ {\small$\pm$0.02} 
                    & 86.72\ {\small$\pm$0.03}
                    & 87.00 
                    & 6.3{\tiny$\!\times\!$10}{\small$^{\text{-}9}$}
                    \\
    Ablate router learning rate scaling
                    & 78.79\ {\small$\pm$0.01} 
                    & 86.81\ {\small$\pm$0.03} 
                    & 86.96
                    & 9.1{\tiny$\!\times\!$10}{\small$^{\text{-}8}$}
                    \\
    Ablate per-sample routing
                    & 79.00\ {\small$\pm$0.04}
                    & 87.11\ {\small$\pm$0.04}
                    & 87.28 
                    & 0.13
                    \\
    Ablate router bias initialization
                    & 79.19\ {\small$\pm$0.04}
                    & 87.12\ {\small$\pm$0.04}
                    & 87.27
                    & 0.22
                    \\
\midrule
\multicolumn{5}{c}{\emph{Multipath agent: Headroom analysis}} \\
    Force \href{https://www.tensorflow.org/datasets/catalog/i_naturalist2017}{i\_naturalist2017} selection
                    &  79.33\ {\small$\pm$0.02} 
                    & 87.26\ {\small$\pm$0.03}
                    & 87.39 
                    & 0.08
                    \\
\bottomrule
\end{tabular}

\end{center}
\end{table*}

\section{Experiments}
\label{section:experiments}

This section reports an empirical study of the proposed multipath agent.
In all the experiments reported, the proposed method is applied to extend the $\mu$3Net system \citep{Gesmundo2022munet4}.
As benchmark we use the \href{https://www.tensorflow.org/datasets/catalog/imagenet2012}{imagenet2012} task \citep{Russakovsky2015ImageNetLS}. This task has been introduced in the $\mu$Net system with a singlepath method by \citet{Gesmundo2022munet2}.
Imagenet2012 has been chosen for the multipath agent study because: 1) it has the biggest headroom compared to state-of-the-art among the 124 tasks solved by the $\mu$3Net system, 2) the models setting the stat-of-the-art for this task are significantly bigger than the singlepath models currently constituting the $\mu$3Net system. Therefore, quality gains can be expected on this benchmark by leveraging more parameters and knowledge.

The ``\emph{repetitions on independent system replicas}'' method (introduced in Section 5.3 of \citet{Gesmundo2022munet2}) is employed to measure the variance of the reported metrics.
Each multipath agent experiment reported is repeated 10 times.
Each repetition is executed on an independent copy of the $\mu$3Net system.
Metrics are reported as average aggregated across the 10 repetitions paired with the standard error of the mean.
Each experiment performs 15 evolutionary cycles.
To accelerate convergence, the search space is restricted by excluding from the set of candidate support paths the models trained on VTAB-1k tasks \citep{Zhai2019TheVT}, as paths trained on corresponding VTAB-full tasks can provide a strictly more informative representation.

All the experiments can be reproduced by using the published artefacts\footnote{Source-code of the multipath agent and checkpoints of different versions of the $\mu$Net system are available at \\ \href{https://github.com/google-research/google-research/tree/master/muNet}{https://github.com/google-research/google-research/tree/master/muNet}}: a) the multipath agent code, b) $\mu$3Net checkpoint, c) tasks from the \href{https://www.tensorflow.org/datasets/catalog/overview}{Tensorflow datesets catalog}.

\begin{figure}[t]
\centering
\text{Methods comparison on the imagenet2012 benchmark}
\par\medskip

\hspace*{-20.82547pt}
\begin{minipage}{.55\textwidth}
  \centering
  \text{\footnotesize Validation accuracy}
  \includegraphics[width=1.0\linewidth]{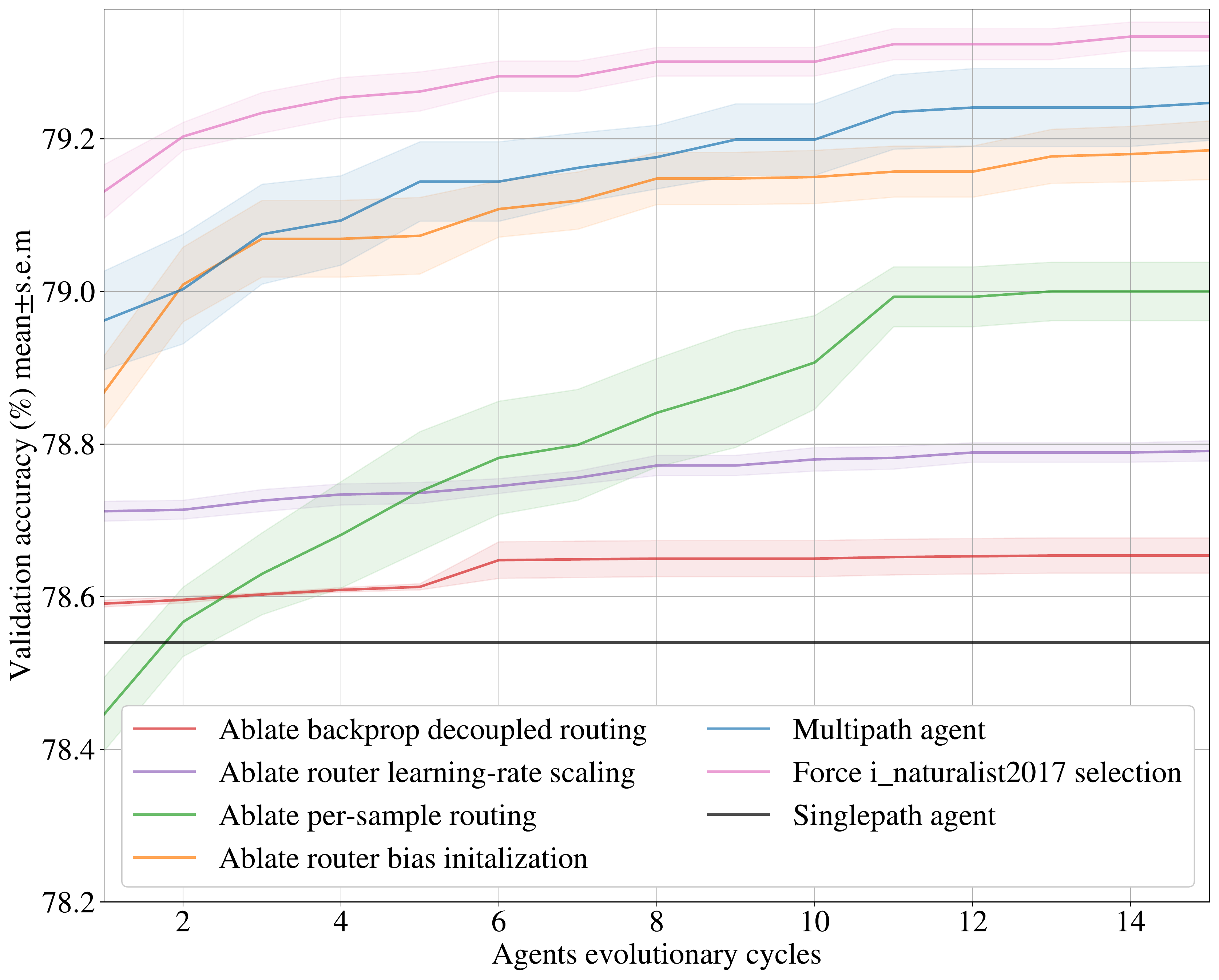}
\end{minipage}%
\begin{minipage}{.55\textwidth}
  \centering
  \text{\footnotesize Test accuracy}
  \includegraphics[width=1.0\linewidth]{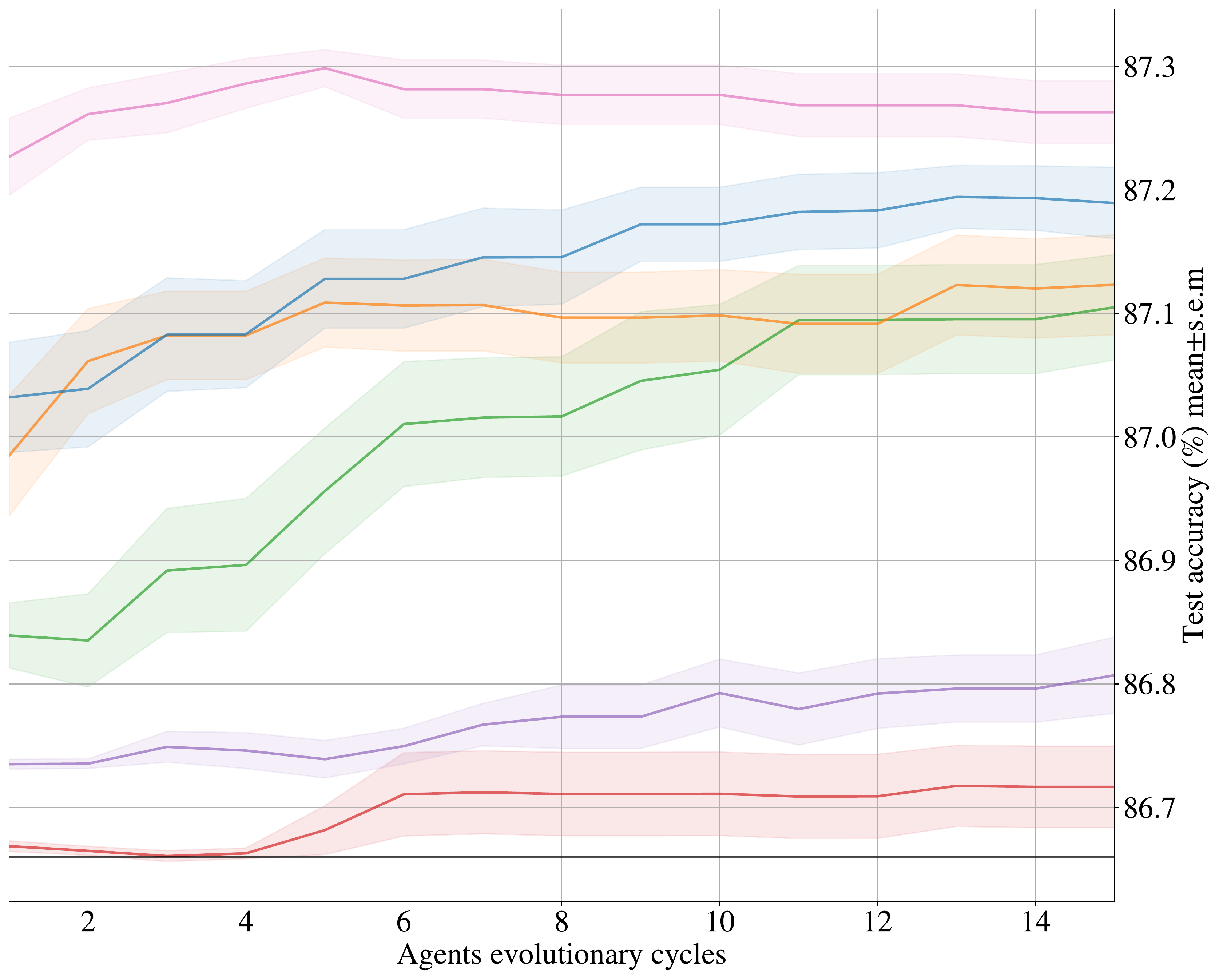}
\end{minipage}
\caption{
Comparison of the validation and test accuracies achieved throughout the evolutionary process by the variations of the multipath agent method described in Section~\ref{section:experiments}.
Each curve represents the accuracy averaged across the 10 experiment repetitions executing the corresponding multipath method variation.
The shaded area represents the standard error of the mean.
The accuracy achieved by the singlepath agent is displayed as an horizontal line for reference.
Numerical values of the accuracies achieved at the completion of the $15^{th}$ evolutionary cycle are reported in Table~\ref{table:bestmetrics}.
Results are discussed in Section~\ref{section:experiments}.
}
\label{fig:curves}
\end{figure}

\subsection{Baseline methods}

The quality achieved by the proposed multipath agent on the imagenet2012 benchmark are reported in Table~\ref{table:bestmetrics} and Figure~\ref{fig:curves}.
The achieved average test accuracy of 87.19\% can be considered significantly higher than the average test accuracy of 85.30\% \citep{Dosovitskiy2021AnII} achieved by fine-tuning a ViT Large model equivalent to the root model used by $\mu$Net system \citep{Gesmundo2022munet2}.

Quality gains are also achieved against the singlepath model for imagenet2012 that is part of the $\mu$3Net system: 86.66\% \citep{Gesmundo2022munet4}.
Notice that this is the path used as \emph{main path} for the architectures generated by the multipath agent.

\subsection{Ablation study}
\label{ssec:ablation}

The proposed multipath agent method is characterize by particular design elements:
\begin{enumerate}
    \item Per-sample routing (see Section~\ref{sec:arch}).
    \item Router bias initialization (see Section~\ref{sec:init_bias}).
    \item Backprop decoupled routing (see Section~\ref{sec:bdr}).
    \item Router learning rate scaling (see Section~\ref{sec:lrm}).
\end{enumerate}
Section~\ref{section:method} introduces the motivating intuitions for each of these design elements.
During the development phase, quality gains have been measured for each design element with preliminary experiments.
In this section, we validate the findings with reproducible experiments.
Each experiment ablates one of the design elements to measure its impact on convergence speed and quality achieved within the training budget of 15 evolutionary cycles.

\subsubsection{Ablate backprop decoupled routing}

The backward propagation decoupled routing method has been introduced as a solution to the \emph{rich~gets~richer} effect (see Section~\ref{sec:bdr}).
In the context of the proposed method, this effect causes the router to early converge to select exclusively the main path, thus preventing the support paths from training.
Therefore, disabling this design element is expected to result in the quality of the multipath models to converge to the quality of the main path in most cases.

This ablation is performed by replacing the \emph{backprop decoupled routing} (equation~\ref{eq:bdr}) the standard router weighted aggregation (equation~\ref{eq:2}).
Note that, the forward pass and gradient computation of the router are unaltered by this ablation, only the backward pass of the connector modules is altered.

Table~\ref{table:bestmetrics} and Figure~\ref{fig:curves} report the results.
8 out of the 10 replicas collapse to selecting exclusively the main path, thus resulting into an test accuracy equivalent to that of the main path: $\in$[86.65\%, 86.68\%].
Only two repetitions avoid the collapse effect and achieve a higher test accuracy: 87.00\% and 86.80\%.

We propose the \emph{backprop decouple routing} solution as it is compatible with the modular nature of the methodology developed by the $\mu$Net line of research (see Section~\ref{sec:bdr}). Such modularity is not provided by the common solutions to \emph{router early convergence} issues such as auxiliary loss factors.

\subsubsection{Ablate per-sample routing}

The router component provides weights determining the influence of each path onto the output. The router is conditioned on a representation of each input sample, thus allowing to provide a different aggregation weighting for each sample.
To evaluate the effect of the per-sample router, we ablate it by replacing the weighted aggregation of with a sum aggregation.
This results in the removal of the router component, and loss of per-sample conditioning of the aggregation.

In practice, this ablation is achieved by replacing the aggregation logic (equation~\ref{eq:bdr}) with a sum of the representation, that can be expressed as: $\mathbf{a} = R \cdot \mathbf{1}$.
Note that, this results in a different forward pass computation, however the gradients computed for the connector modules are unaltered.

This ablation results in lower average accuracy.
The quality loss is more pronounced in the first half of the experiment and on the validation set.


Notice that, the proposed method requires to compute all the representations of the paths included in the sampled architecture as the router function is limited to produce aggregation weights (\emph{soft-routing}). This method can be extended to sparsely activate the alternative paths conditioning on the router weights (\emph{hard routing}) as it is usual in sparsely activated mixture-of-expert layers. Hard routing extensions can allow to achieve higher compute efficiency, scaling or per-sample erogenous compute properties. In such extensions the role of the per-sample router module becomes critical as the selection of the paths to activate depends on its output.

\subsubsection{Ablate router bias initialization}

The \emph{router bias initialization} is ablated by applying zero initialization to all the parameters of the router.
This leads the initialized router to produce a uniform distribution of routing weights.
This ablation has the weaker significance compared to the other considered.
In preliminary experiments, we observed a stronger significance on tasks with smaller training set, as the initial bias becomes less relevant the longer the training.

\subsubsection{Ablate router learning rate scaling}
The \emph{learning rate scaling} technique results having a critical impact on the achieved quality.
The learning rate scaling is ablated by simply applying the global learning rate also to the router component, i.e. disabling the logic represented by equation~(\ref{eq:lrs2}).

\subsection{Headroom analysis}
\label{ssec:headroom}
The set of paths connected by a multitask architecture ($\mathcal{P}$) is sampled at random among the set of all the paths available in the system: $\mathcal{S}$.
Even though the configuration of the reported experiments caps the maximum amount of paths to 3, the resulting search space of paths combinations cannot be exhaustively explored with the provided exploratory budget of 15 evolutionary cycles (16 models are sampled each cycle).
Dynamic systems such as $\mu$Net are designed to continuously expand to solve an unbounded number of tasks: $|\mathcal{S}|\rightarrow\inf$.
In this context, the random sampling method is expected to decrease in efficacy as the number of possible paths combination grows rapidly: number of paths combinations ${=|\mathcal{S}|!\ /\ ((|\mathcal{S}|\!-\!|\mathcal{P}|)! \cdot |\mathcal{P}|!)}$.

The experiment reported in this section aims to estimate the headroom available for improved agents capable of performing a more informed selection of the paths that contain relevant knowledge for each task/sample.
This experiment executes the proposed multipath agent with a constraint in the selection of the first support path.
The path trained on i\_naturalist2017 is forced to be used as first support path.
This path has been chosen based on an analysis done on the best scoring multipath architectures generated by the multipath experiments reported in previous sections.
The analysis shows that a small set of paths are frequently selected as support in the highest scoring models, these are the paths are trained on tasks such as: \href{https://www.tensorflow.org/datasets/catalog/i_naturalist2017}{i\_naturalist2017},
\href{https://www.tensorflow.org/datasets/catalog/imagenet_a}{imagenet\_a},
\href{https://www.tensorflow.org/datasets/catalog/imagenet_r}{imagenet\_r}.
imagenet\_a and
imagenet\_r are datasets have explicitly collected to augment the imagenet2012 task.
While i\_naturalist2017 is a large dataset of natural domain images. More than 30\% of imagenet2012 classes identify entities in the natural domain.
Furthermore, these paths carry knowledge of multiple tasks on which their ancestors have been trained. For example, imagenet\_a appears to have been mutated from i\_naturalist2017 and still share most of its components (see Figure~\ref{fig:arc-large}).
Forcing the assignment of the first support path has the effect of restricting the search space to a size that is more feasibly explorable with random sampling within the given evolutionary budget. And may give us a sense of what is the headroom for methods that use a more informed logic to select the activated paths (e.g. condition the sampling distribution on paths/task meta-features or on the results achieved by past samplings).

The result of this experiment show that: 1) higher quality is achievable, 2) the peak test accuracy (within noise) can be achieved with a significant reduced exploratory budget.


\section{Related work}

This paper contributes to a line of research aiming to define and demonstrate a novel ML research methodology enabling the creation of \emph{dynamic large-scale multi-modal multitask modular intelligent systems} that can be indefinitely extended through the collaboration of multiple contributors: \citep{Gesmundo2022munet1,Gesmundo2022munet2,Gesmundo2022munet3,Gesmundo2022munet4}.

In the following paragraphs, we refer to prior work that has been influential in the design of the multipath agent method.

\textbf{Ensemble learning} relies on the parallel combination of several models to obtain better generalization \citep{Kurutach2018ModelEnsembleTP}.
This technique has been applied consistently with success also to neural networks, providing quality gains and higher training stability in the vast majority of its applications.
The success of the proposed parallel activation of paths can be interpreted with similar intuitions to those motivating ensemble learning. Although, there are major differences: 1) models ensembling combines models trained on the same tasks, while the proposed method combines model/paths trained on different tasks, 2) models ensembling allows to compose frozen models while the aggregation logic can be subject to training independently, instead the multipath activation requires training of the paths trained on different tasks to adapt to the target task (i.e. with trainable connector modules), 3) applications of ensemble learning do not normally include per-sample conditioned aggregation.

Per-sample \textbf{routing} is a characteristic of mixture-of-expert (MoE) methods \citep{Shazeer2017OutrageouslyLN}. MoE routers are conditioned on the same input provided to the experts. While, the proposed design leverages the presence of a \emph{main path}. This allows to provide a more informative representation of the input to the router, that enables to minimize the number of trainable parameters (i.e. the router is a single fully connected layer). Furthermore we propose alternative solutions to the common routing convergence issues that are compatible with the modular nature of the proposed research methodology.

Prior work has also explored the application of \textbf{distinct learning rates} to different components of a sparsely activated architecture \citep{Artetxe2021EfficientLS,Fedus2022ARO}.

\emph{Backprop decoupled routing} is a technique that relies on the decoupling of the forward and backward pass logic. Although defined to provide a different functionality, the \textbf{gumbel-softmax} technique \citep{Jang2016CategoricalRW} also relies on decoupling of the forward and backward pass.

\citet{Merullo2022LinearlyMF} has demonstrated the possibility to achieve competitive performance by connecting frozen models with \textbf{trainable connectors}.
Their work focuses on serial connections, while the multipath architecture applies a similar intuition to parallel connections. Also, both methods rely on the use of a architecturally minimalistic connector.

\subsection{Extended survey}
In the remainder of this section, we provide a survey of topics related to the $\mu$Net line of research.

The proposed system is designed to be immune from common multitask learning pitfalls: catastrophic forgetting, gradients interference, negative transfer. 
Cross-task \textbf{transfer-learning}
has gained popularity, especially through transfer learning from a model
pre-trained on a large amount of data for one or a few general tasks,
and then fine-tuned on a small amount of data for a related downstream task.
This approach has been shown to be very effective in a wide variety of problems
across many modalities, including
language \citep{Devlin2019BERTPO,Raffel2020ExploringTL} and vision \citep{Dosovitskiy2021AnII,He2016DeepRL}.
The success of transfer-learning applications hinges on adequate prior knowledge selection to avoid typical \textbf{negative transfer} pitfalls \citep{Rosenstein2005ToTO,Wang2019CharacterizingAA}.
Common solutions rely on data or model selection techniques,
often putting emphasis on the efficiency of the exploration
\citep{Zhang2020ASO,Mensink2021FactorsOI}, also method aiming to automate knowledge selection at a layer level have been proposed \citet{Sun2020AdaShareLW}.
Transfer learning capabilities are critical for \textbf{multitask models}.
ML models trained jointly on multiple tasks 
can be affected by \textbf{gradients interference} if any subset of parameters receive gradients jointly from multiple sources \citep{Chen2018GradNormGN,Yu2020GradientSF}, and by \textbf{catastrophic forgetting} of prior knowledge as new tasks are learned \citep{McCloskey1989CatastrophicII,French1999CatastrophicFI}.
These knowledge loss problems can be alleviated with weighted combination of tasks \citep{Liu2019LossBalancedTW,Sun2020ERNIE2A} and gradient transformation methods \citep{Chen2018GradNormGN,Sener2018MultiTaskLA,Kendall2018MultitaskLU}.
Stronger guarantees are provided by methods that compartmentalize task specific knowledge in dedicated parameter subsets \citep{Rebuffi2017LearningMV,Houlsby2019ParameterEfficientTL,Rusu2016ProgressiveNN,Rosenfeld2020IncrementalLT}.
Addressing catastrophic forgetting and identifying what subset of parameters/knowledge that is beneficial to share with each task is also critical for \textbf{continual learning} or life long learning methods
\citep{McCloskey1989CatastrophicII,French1999CatastrophicFI,Ramesh2022ModelZA}, some also considering a modular approach \citep{Vniat2020EfficientCL}.

The proposed method relies on an evolutionary approach to jointly search the spaces of models architectures, hyperparameters, and prior knowledge selection.
The automation of \textbf{hyperparameter tuning} has been commonly addressed with Bayesian optimization \citep{Srinivas2010GaussianPO,Bergstra2011AlgorithmsFH,Snoek2012PracticalBO},
evolutionary methods have also been explored for this purpose
\citep{Jaderberg2017PopulationBT,Zhang2011EvolutionaryCM}.
Hyperparameters tuning can be considered related to the \textbf{neural architecture search} (NAS), as architectures can be defined by the selection of a sequence
of architectural hyperparameters.
Initially, NAS methods have been based on reinforcement learning techniques
\citep{Zoph2017NeuralAS} but also sample efficient evolutionary approaches have been proposed \citep{Real2019RegularizedEF,Maziarz2018EvolutionaryNeuralHA}.
Parameter-sharing based NAS methods aim to reduce the typically high training cost \citep{Pham2018EfficientNA,Liu2019DARTSDA,Kokiopoulou2019FastTA}.
Optimization for multi-factor quality/cost trade-offs have been explored \citep{Tan2019MnasNetPN}.

Prior methods have been proposed to achieve \textbf{dynamic architecture extensions} \citep{Chen2016Net2NetAL,Cai2018EfficientAS}, some also focusing on an unbounded stream of tasks \citep{Yoon2018LifelongLW,Yao2020OnlineSM}, or achieving immunity from catastrophic forgetting  \citep{Rusu2016ProgressiveNN,Li2018LearningWF,Li2019LearnTG,Rosenfeld2020IncrementalLT}.

Sparse activation allows to decouple knowledge/parameters growth from computational cost increase.
Performance improvements of state-of-the-art models often requires growth in terms of trainable parameters \citep{Kaplan2020ScalingLF}.
\textbf{Sparse activation} techniques at sub-layer level \citep{Shazeer2017OutrageouslyLN,Du2021GLaMES} or network route level \citep{Fernando2017PathNetEC} allow to decouple model size growth from compute cost.
This is achieved by integrating
a \textbf{routing technique} that selects the appropriate subset of parameters storing the most relevant knowledge for each task, sample or token/patch.

The ability of jointly solve a \textbf{large amount of tasks} is commonly associated with progress toward Artificial General Intelligence (AGI).
Advancements in scaling language models \citep{Brown2020LanguageMA,Thoppilan2022LaMDALM} allowed to achieve novel discourse, reasoning and zero/few shot learning capabilities that can be applied to new tasks without/minimal additional training.
Recent work aims to extend these achievements beyond text modality by defining static architectures for an extended subset of modalities \citep{Alayrac2022FlamingoAV,Reed2022AGA}.
These are a few examples of the ML models contributing to the line of research achieving incremental milestone toward AGI.
Though, each model is trained from scratch with considerable resources consumption.
The introduction of abstractions allowing to modularize, dynamically extend and reuse these large models may contribute to accelerate the rate of innovation.

Large scale $\mu$Net experiments have been enabled by the use of the ML-Pathways framework \citep{Barham2022PathwaysAD}.

\section{Conclusion}

We presented a novel agent capable of generating architectures activating multiple paths within a modular multitask system such as $\mu$Net.
Furthermore, paths activation is conditioned on the input allowing to select and combine the relevant knowledge for each sample. 
We demonstrated that such method can be applied to achieve the quality headroom available to tasks with high ambiguity and large dataset that can benefit from the use of models with more knowledge/parameters such as imagenet2012.
We defined novel design elements of critical importance, such as: \emph{modular learning rate scaling} and \emph{backprop decoupled routing}. These elements are compatible with the modular nature of the novel methodology developed by the $\mu$Net line of research, but are also generically applicable to arbitrary neural network architectures.

We have provided empirical evidences suggesting that further headroom can be achieved with a mutation sampling logic of higher sophistication than random sampling.
We expect this headroom to widen as the complexity of the design space increases with more tasks and methods being added to continuously expanding systems such as $\mu$Net.

Future work can continue to build toward extending the set of tasks and connectable model architectures and enabling transfer of knowledge across different input/output modalities.


\bibliography{iclr2023_conference}

\begin{thebibliography}{62}
\providecommand{\natexlab}[1]{#1}
\providecommand{\url}[1]{\texttt{#1}}
\expandafter\ifx\csname urlstyle\endcsname\relax
  \providecommand{\doi}[1]{doi: #1}\else
  \providecommand{\doi}{doi: \begingroup \urlstyle{rm}\Url}\fi

\bibitem[Alayrac et~al.(2022)Alayrac, Donahue, Luc, Miech, Barr, Hasson, Lenc,
  Mensch, Millican, Reynolds, Ring, Rutherford, Cabi, Han, Gong, Samangooei,
  Monteiro, Menick, Borgeaud, Brock, Nematzadeh, Sharifzadeh, Binkowski,
  Barreira, Vinyals, Zisserman, and Simonyan]{Alayrac2022FlamingoAV}
Jean-Baptiste Alayrac, Jeff Donahue, Pauline Luc, Antoine Miech, Iain Barr,
  Yana Hasson, Karel Lenc, Arthur Mensch, Katie Millican, Malcolm Reynolds,
  Roman Ring, Eliza Rutherford, Serkan Cabi, Tengda Han, Zhitao Gong, Sina
  Samangooei, Marianne Monteiro, Jacob Menick, Sebastian Borgeaud, Andy Brock,
  Aida Nematzadeh, Sahand Sharifzadeh, Mikolaj Binkowski, Ricardo Barreira,
  Oriol Vinyals, Andrew Zisserman, and Karen Simonyan.
\newblock Flamingo: a visual language model for few-shot learning.
\newblock \emph{ArXiv}, abs/2204.14198, 2022.

\bibitem[Artetxe et~al.(2021)Artetxe, Bhosale, Goyal, Mihaylov, Ott, Shleifer,
  Lin, Du, Iyer, Pasunuru, Anantharaman, Li, Chen, Akın, Baines, Martin, Zhou,
  Koura, O'Horo, Wang, Zettlemoyer, Diab, Kozareva, and
  Stoyanov]{Artetxe2021EfficientLS}
Mikel Artetxe, Shruti Bhosale, Naman Goyal, Todor Mihaylov, Myle Ott, Sam
  Shleifer, Xi~Victoria Lin, Jingfei Du, Srini Iyer, Ramakanth Pasunuru,
  Giridhar Anantharaman, Xian Li, Shuohui Chen, Halil Akın, Mandeep Baines,
  Louis Martin, Xing Zhou, Punit~Singh Koura, Brian O'Horo, Jeff Wang, Luke
  Zettlemoyer, Mona Diab, Zornitsa Kozareva, and Ves Stoyanov.
\newblock Efficient large scale language modeling with mixtures of experts.
\newblock \emph{ArXiv}, abs/2112.10684, 2021.

\bibitem[Barham et~al.(2022)Barham, Chowdhery, Dean, Ghemawat, Hand, Hurt,
  Isard, Lim, Pang, Roy, Saeta, Schuh, Sepassi, Shafey, Thekkath, and
  Wu]{Barham2022PathwaysAD}
Paul Barham, Aakanksha Chowdhery, Jeffrey Dean, Sanjay Ghemawat, Steven Hand,
  Daniel Hurt, Michael Isard, Hyeontaek Lim, Ruoming Pang, Sudip Roy, Brennan
  Saeta, Parker Schuh, Ryan Sepassi, Laurent~El Shafey, Chandramohan~A.
  Thekkath, and Yonghui Wu.
\newblock Pathways: Asynchronous distributed dataflow for ml.
\newblock \emph{ArXiv}, abs/2203.12533, 2022.

\bibitem[Bergstra et~al.(2011)Bergstra, Bardenet, Bengio, and
  K{\'e}gl]{Bergstra2011AlgorithmsFH}
James Bergstra, R{\'e}mi Bardenet, Yoshua Bengio, and Bal{\'a}zs K{\'e}gl.
\newblock Algorithms for hyper-parameter optimization.
\newblock In \emph{NIPS}, 2011.

\bibitem[Brown et~al.(2020)Brown, Mann, Ryder, Subbiah, Kaplan, Dhariwal,
  Neelakantan, Shyam, Sastry, Askell, Agarwal, Herbert-Voss, Krueger, Henighan,
  Child, Ramesh, Ziegler, Wu, Winter, Hesse, Chen, Sigler, Litwin, Gray, Chess,
  Clark, Berner, McCandlish, Radford, Sutskever, and
  Amodei]{Brown2020LanguageMA}
Tom~B. Brown, Benjamin Mann, Nick Ryder, Melanie Subbiah, Jared Kaplan,
  Prafulla Dhariwal, Arvind Neelakantan, Pranav Shyam, Girish Sastry, Amanda
  Askell, Sandhini Agarwal, Ariel Herbert-Voss, Gretchen Krueger, T.~J.
  Henighan, Rewon Child, Aditya Ramesh, Daniel~M. Ziegler, Jeff Wu, Clemens
  Winter, Christopher Hesse, Mark Chen, Eric Sigler, Mateusz Litwin, Scott
  Gray, Benjamin Chess, Jack Clark, Christopher Berner, Sam McCandlish, Alec
  Radford, Ilya Sutskever, and Dario Amodei.
\newblock Language models are few-shot learners.
\newblock \emph{ArXiv}, abs/2005.14165, 2020.

\bibitem[Cai et~al.(2018)Cai, Chen, Zhang, Yu, and Wang]{Cai2018EfficientAS}
Han Cai, Tianyao Chen, Weinan Zhang, Yong Yu, and Jun Wang.
\newblock Efficient architecture search by network transformation.
\newblock In \emph{AAAI}, 2018.

\bibitem[Chen et~al.(2016)Chen, Goodfellow, and Shlens]{Chen2016Net2NetAL}
Tianqi Chen, Ian~J. Goodfellow, and Jonathon Shlens.
\newblock Net2net: Accelerating learning via knowledge transfer.
\newblock \emph{CoRR}, abs/1511.05641, 2016.

\bibitem[Chen et~al.(2018)Chen, Badrinarayanan, Lee, and
  Rabinovich]{Chen2018GradNormGN}
Zhao Chen, Vijay Badrinarayanan, Chen-Yu Lee, and Andrew Rabinovich.
\newblock Gradnorm: Gradient normalization for adaptive loss balancing in deep
  multitask networks.
\newblock In \emph{ICML}, 2018.

\bibitem[Devlin et~al.(2019)Devlin, Chang, Lee, and
  Toutanova]{Devlin2019BERTPO}
Jacob Devlin, Ming-Wei Chang, Kenton Lee, and Kristina Toutanova.
\newblock Bert: Pre-training of deep bidirectional transformers for language
  understanding.
\newblock In \emph{NAACL}, 2019.

\bibitem[Dosovitskiy et~al.(2021)Dosovitskiy, Beyer, Kolesnikov, Weissenborn,
  Zhai, Unterthiner, Dehghani, Minderer, Heigold, Gelly, Uszkoreit, and
  Houlsby]{Dosovitskiy2021AnII}
Alexey Dosovitskiy, Lucas Beyer, Alexander Kolesnikov, Dirk Weissenborn,
  Xiaohua Zhai, Thomas Unterthiner, Mostafa Dehghani, Matthias Minderer, Georg
  Heigold, Sylvain Gelly, Jakob Uszkoreit, and Neil Houlsby.
\newblock An image is worth 16x16 words: Transformers for image recognition at
  scale.
\newblock \emph{ArXiv}, abs/2010.11929, 2021.

\bibitem[Du et~al.(2021)Du, Huang, Dai, Tong, Lepikhin, Xu, Krikun, Zhou, Yu,
  Firat, Zoph, Fedus, Bosma, Zhou, Wang, Wang, Webster, Pellat, Robinson,
  Meier-Hellstern, Duke, Dixon, Zhang, Le, Wu, Chen, and Cui]{Du2021GLaMES}
Nan Du, Yanping Huang, Andrew~M. Dai, Simon Tong, Dmitry Lepikhin, Yuanzhong
  Xu, Maxim Krikun, Yanqi Zhou, Adams~Wei Yu, Orhan Firat, Barret Zoph, Liam
  Fedus, Maarten Bosma, Zongwei Zhou, Tao Wang, Yu~Emma Wang, Kellie Webster,
  Marie Pellat, Kevin Robinson, Kathleen~S. Meier-Hellstern, Toju Duke, Lucas
  Dixon, Kun Zhang, Quoc~V. Le, Yonghui Wu, Z.~Chen, and Claire Cui.
\newblock Glam: Efficient scaling of language models with mixture-of-experts.
\newblock \emph{ArXiv}, abs/2112.06905, 2021.

\bibitem[Fedus et~al.(2022)Fedus, Dean, and Zoph]{Fedus2022ARO}
William Fedus, Jeff Dean, and Barret Zoph.
\newblock A review of sparse expert models in deep learning.
\newblock \emph{ArXiv}, abs/2209.01667, 2022.

\bibitem[Fernando et~al.(2017)Fernando, Banarse, Blundell, Zwols, Ha, Rusu,
  Pritzel, and Wierstra]{Fernando2017PathNetEC}
Chrisantha Fernando, Dylan~S. Banarse, Charles Blundell, Yori Zwols, David~R
  Ha, Andrei~A. Rusu, Alexander Pritzel, and Daan Wierstra.
\newblock Pathnet: Evolution channels gradient descent in super neural
  networks.
\newblock \emph{ArXiv}, abs/1701.08734, 2017.

\bibitem[French(1999)]{French1999CatastrophicFI}
Robert~M. French.
\newblock Catastrophic forgetting in connectionist networks.
\newblock \emph{Trends in Cognitive Sciences}, 3:\penalty0 128--135, 1999.

\bibitem[Gesmundo(2022{\natexlab{a}})]{Gesmundo2022munet3}
Andrea Gesmundo.
\newblock A continual development methodology for large-scale multitask dynamic
  ml systems.
\newblock \emph{ArXiv}, 2209.07326, 2022{\natexlab{a}}.

\bibitem[Gesmundo(2022{\natexlab{b}})]{Gesmundo2022munet4}
Andrea Gesmundo.
\newblock A multiagent framework for the asynchronous and collaborative
  extension of multitask ml systems.
\newblock \emph{ArXiv}, 2209.14745, 2022{\natexlab{b}}.

\bibitem[Gesmundo \& Dean(2022{\natexlab{a}})Gesmundo and
  Dean]{Gesmundo2022munet1}
Andrea Gesmundo and Jeff Dean.
\newblock munet: Evolving pretrained deep neural networks into scalable
  auto-tuning multitask systems.
\newblock \emph{ArXiv}, 2205.10937, 2022{\natexlab{a}}.

\bibitem[Gesmundo \& Dean(2022{\natexlab{b}})Gesmundo and
  Dean]{Gesmundo2022munet2}
Andrea Gesmundo and Jeff Dean.
\newblock An evolutionary approach to dynamic introduction of tasks in
  large-scale multitask learning systems.
\newblock \emph{ArXiv}, 2205.12755, 2022{\natexlab{b}}.

\bibitem[He et~al.(2016)He, Zhang, Ren, and Sun]{He2016DeepRL}
Kaiming He, X.~Zhang, Shaoqing Ren, and Jian Sun.
\newblock Deep residual learning for image recognition.
\newblock \emph{2016 IEEE Conference on Computer Vision and Pattern Recognition
  (CVPR)}, pp.\  770--778, 2016.

\bibitem[Houlsby et~al.(2019)Houlsby, Giurgiu, Jastrzebski, Morrone,
  de~Laroussilhe, Gesmundo, Attariyan, and
  Gelly]{Houlsby2019ParameterEfficientTL}
Neil Houlsby, Andrei Giurgiu, Stanislaw Jastrzebski, Bruna Morrone, Quentin
  de~Laroussilhe, Andrea Gesmundo, Mona Attariyan, and Sylvain Gelly.
\newblock Parameter-efficient transfer learning for nlp.
\newblock In \emph{ICML}, 2019.

\bibitem[Jaderberg et~al.(2017)Jaderberg, Dalibard, Osindero, Czarnecki,
  Donahue, Razavi, Vinyals, Green, Dunning, Simonyan, Fernando, and
  Kavukcuoglu]{Jaderberg2017PopulationBT}
Max Jaderberg, Valentin Dalibard, Simon Osindero, Wojciech~M. Czarnecki, Jeff
  Donahue, Ali Razavi, Oriol Vinyals, Tim Green, Iain Dunning, Karen Simonyan,
  Chrisantha Fernando, and Koray Kavukcuoglu.
\newblock Population based training of neural networks.
\newblock \emph{ArXiv}, abs/1711.09846, 2017.

\bibitem[Jang et~al.(2016)Jang, Gu, and Poole]{Jang2016CategoricalRW}
Eric Jang, Shixiang~Shane Gu, and Ben Poole.
\newblock Categorical reparameterization with gumbel-softmax.
\newblock \emph{ArXiv}, abs/1611.01144, 2016.

\bibitem[Kaplan et~al.(2020)Kaplan, McCandlish, Henighan, Brown, Chess, Child,
  Gray, Radford, Wu, and Amodei]{Kaplan2020ScalingLF}
Jared Kaplan, Sam McCandlish, T.~J. Henighan, Tom~B. Brown, Benjamin Chess,
  Rewon Child, Scott Gray, Alec Radford, Jeff Wu, and Dario Amodei.
\newblock Scaling laws for neural language models.
\newblock \emph{ArXiv}, abs/2001.08361, 2020.

\bibitem[Kendall et~al.(2018)Kendall, Gal, and Cipolla]{Kendall2018MultitaskLU}
Alex Kendall, Yarin Gal, and Roberto Cipolla.
\newblock Multi-task learning using uncertainty to weigh losses for scene
  geometry and semantics.
\newblock \emph{2018 IEEE/CVF Conference on Computer Vision and Pattern
  Recognition}, pp.\  7482--7491, 2018.

\bibitem[Kokiopoulou et~al.(2019)Kokiopoulou, Hauth, Sbaiz, Gesmundo,
  Bart{\'o}k, and Berent]{Kokiopoulou2019FastTA}
Efi Kokiopoulou, Anja Hauth, Luciano Sbaiz, Andrea Gesmundo, G{\'a}bor
  Bart{\'o}k, and Jesse Berent.
\newblock Fast task-aware architecture inference.
\newblock \emph{ArXiv}, abs/1902.05781, 2019.

\bibitem[Kurutach et~al.(2018)Kurutach, Clavera, Duan, Tamar, and
  Abbeel]{Kurutach2018ModelEnsembleTP}
Thanard Kurutach, Ignasi Clavera, Yan Duan, Aviv Tamar, and P.~Abbeel.
\newblock Model-ensemble trust-region policy optimization.
\newblock \emph{ArXiv}, abs/1802.10592, 2018.

\bibitem[Li et~al.(2019)Li, Zhou, Wu, Socher, and Xiong]{Li2019LearnTG}
Xilai Li, Yingbo Zhou, Tianfu Wu, Richard Socher, and Caiming Xiong.
\newblock Learn to grow: A continual structure learning framework for
  overcoming catastrophic forgetting.
\newblock In \emph{ICML}, 2019.

\bibitem[Li \& Hoiem(2018)Li and Hoiem]{Li2018LearningWF}
Zhizhong Li and Derek Hoiem.
\newblock Learning without forgetting.
\newblock \emph{IEEE Transactions on Pattern Analysis and Machine
  Intelligence}, 40:\penalty0 2935--2947, 2018.

\bibitem[Liu et~al.(2019{\natexlab{a}})Liu, Simonyan, and Yang]{Liu2019DARTSDA}
Hanxiao Liu, Karen Simonyan, and Yiming Yang.
\newblock Darts: Differentiable architecture search.
\newblock \emph{ArXiv}, abs/1806.09055, 2019{\natexlab{a}}.

\bibitem[Liu et~al.(2019{\natexlab{b}})Liu, Liang, and
  Gitter]{Liu2019LossBalancedTW}
Shengchao Liu, Yingyu Liang, and Anthony Gitter.
\newblock Loss-balanced task weighting to reduce negative transfer in
  multi-task learning.
\newblock In \emph{AAAI}, 2019{\natexlab{b}}.

\bibitem[Maziarz et~al.(2018)Maziarz, Khorlin, de~Laroussilhe, and
  Gesmundo]{Maziarz2018EvolutionaryNeuralHA}
Krzysztof Maziarz, Andrey Khorlin, Quentin de~Laroussilhe, and Andrea Gesmundo.
\newblock Evolutionary-neural hybrid agents for architecture search.
\newblock \emph{ArXiv}, abs/1811.09828, 2018.

\bibitem[McCloskey \& Cohen(1989)McCloskey and
  Cohen]{McCloskey1989CatastrophicII}
Michael McCloskey and Neal~J. Cohen.
\newblock Catastrophic interference in connectionist networks: The sequential
  learning problem.
\newblock \emph{Psychology of Learning and Motivation}, 24:\penalty0 109--165,
  1989.

\bibitem[Mensink et~al.(2021)Mensink, Uijlings, Kuznetsova, Gygli, and
  Ferrari]{Mensink2021FactorsOI}
Thomas Mensink, Jasper R.~R. Uijlings, Alina Kuznetsova, Michael Gygli, and
  Vittorio Ferrari.
\newblock Factors of influence for transfer learning across diverse appearance
  domains and task types.
\newblock \emph{IEEE transactions on pattern analysis and machine
  intelligence}, PP, 2021.

\bibitem[Merullo et~al.(2022)Merullo, Castricato, Eickhoff, and
  Pavlick]{Merullo2022LinearlyMF}
Jack Merullo, Louis Castricato, Carsten Eickhoff, and Elizabeth-Jane Pavlick.
\newblock Linearly mapping from image to text space.
\newblock \emph{ArXiv}, abs/2209.15162, 2022.

\bibitem[Pham et~al.(2018)Pham, Guan, Zoph, Le, and Dean]{Pham2018EfficientNA}
Hieu Pham, Melody~Y. Guan, Barret Zoph, Quoc~V. Le, and Jeff Dean.
\newblock Efficient neural architecture search via parameter sharing.
\newblock In \emph{ICML}, 2018.

\bibitem[Raffel et~al.(2020)Raffel, Shazeer, Roberts, Lee, Narang, Matena,
  Zhou, Li, and Liu]{Raffel2020ExploringTL}
Colin Raffel, Noam~M. Shazeer, Adam Roberts, Katherine Lee, Sharan Narang,
  Michael Matena, Yanqi Zhou, Wei Li, and Peter~J. Liu.
\newblock Exploring the limits of transfer learning with a unified text-to-text
  transformer.
\newblock \emph{ArXiv}, abs/1910.10683, 2020.

\bibitem[Ramesh \& Chaudhari(2022)Ramesh and Chaudhari]{Ramesh2022ModelZA}
Rahul Ramesh and Pratik Chaudhari.
\newblock Model zoo: A growing brain that learns continually.
\newblock In \emph{ICLR}, 2022.

\bibitem[Real et~al.(2019)Real, Aggarwal, Huang, and Le]{Real2019RegularizedEF}
Esteban Real, Alok Aggarwal, Yanping Huang, and Quoc~V. Le.
\newblock Regularized evolution for image classifier architecture search.
\newblock In \emph{AAAI}, 2019.

\bibitem[Rebuffi et~al.(2017)Rebuffi, Bilen, and
  Vedaldi]{Rebuffi2017LearningMV}
Sylvestre-Alvise Rebuffi, Hakan Bilen, and Andrea Vedaldi.
\newblock Learning multiple visual domains with residual adapters.
\newblock In \emph{NIPS}, 2017.

\bibitem[Reed et~al.(2022)Reed, Zolna, Parisotto, Colmenarejo, Novikov,
  Barth-Maron, Gimenez, Sulsky, Kay, Springenberg, Eccles, Bruce, Razavi,
  Edwards, Heess, Chen, Hadsell, Vinyals, Bordbar, and de~Freitas]{Reed2022AGA}
Scott Reed, Konrad Zolna, Emilio Parisotto, Sergio~Gomez Colmenarejo, Alexander
  Novikov, Gabriel Barth-Maron, Mai Gimenez, Yury Sulsky, Jackie Kay,
  Jost~Tobias Springenberg, Tom Eccles, Jake Bruce, Ali Razavi, Ashley~D.
  Edwards, Nicolas Manfred~Otto Heess, Yutian Chen, Raia Hadsell, Oriol
  Vinyals, Mahyar Bordbar, and Nando de~Freitas.
\newblock A generalist agent.
\newblock \emph{ArXiv}, abs/2205.06175, 2022.

\bibitem[Rosenfeld \& Tsotsos(2020)Rosenfeld and
  Tsotsos]{Rosenfeld2020IncrementalLT}
Amir Rosenfeld and John~K. Tsotsos.
\newblock Incremental learning through deep adaptation.
\newblock \emph{IEEE Transactions on Pattern Analysis and Machine
  Intelligence}, 42:\penalty0 651--663, 2020.

\bibitem[Rosenstein(2005)]{Rosenstein2005ToTO}
Michael~T. Rosenstein.
\newblock To transfer or not to transfer.
\newblock In \emph{NIPS 2005}, 2005.

\bibitem[Russakovsky et~al.(2015)Russakovsky, Deng, Su, Krause, Satheesh, Ma,
  Huang, Karpathy, Khosla, Bernstein, Berg, and
  Fei-Fei]{Russakovsky2015ImageNetLS}
Olga Russakovsky, Jia Deng, Hao Su, Jonathan Krause, Sanjeev Satheesh, Sean Ma,
  Zhiheng Huang, Andrej Karpathy, Aditya Khosla, Michael~S. Bernstein,
  Alexander~C. Berg, and Li~Fei-Fei.
\newblock Imagenet large scale visual recognition challenge.
\newblock \emph{International Journal of Computer Vision}, 115:\penalty0
  211--252, 2015.

\bibitem[Rusu et~al.(2016)Rusu, Rabinowitz, Desjardins, Soyer, Kirkpatrick,
  Kavukcuoglu, Pascanu, and Hadsell]{Rusu2016ProgressiveNN}
Andrei~A. Rusu, Neil~C. Rabinowitz, Guillaume Desjardins, Hubert Soyer, James
  Kirkpatrick, Koray Kavukcuoglu, Razvan Pascanu, and Raia Hadsell.
\newblock Progressive neural networks.
\newblock \emph{ArXiv}, abs/1606.04671, 2016.

\bibitem[Sener \& Koltun(2018)Sener and Koltun]{Sener2018MultiTaskLA}
Ozan Sener and Vladlen Koltun.
\newblock Multi-task learning as multi-objective optimization.
\newblock In \emph{NeurIPS}, 2018.

\bibitem[Shazeer et~al.(2017)Shazeer, Mirhoseini, Maziarz, Davis, Le, Hinton,
  and Dean]{Shazeer2017OutrageouslyLN}
Noam~M. Shazeer, Azalia Mirhoseini, Krzysztof Maziarz, Andy Davis, Quoc~V. Le,
  Geoffrey~E. Hinton, and Jeff Dean.
\newblock Outrageously large neural networks: The sparsely-gated
  mixture-of-experts layer.
\newblock \emph{ArXiv}, abs/1701.06538, 2017.

\bibitem[Shen et~al.(2019)Shen, Ott, Auli, and Ranzato]{Shen2019MixtureMF}
Tianxiao Shen, Myle Ott, Michael Auli, and Marc'Aurelio Ranzato.
\newblock Mixture models for diverse machine translation: Tricks of the trade.
\newblock In \emph{International Conference on Machine Learning}, 2019.

\bibitem[Snoek et~al.(2012)Snoek, Larochelle, and Adams]{Snoek2012PracticalBO}
Jasper Snoek, H.~Larochelle, and Ryan~P. Adams.
\newblock Practical bayesian optimization of machine learning algorithms.
\newblock In \emph{NIPS}, 2012.

\bibitem[Srinivas et~al.(2010)Srinivas, Krause, Kakade, and
  Seeger]{Srinivas2010GaussianPO}
Niranjan Srinivas, Andreas Krause, Sham~M. Kakade, and Matthias~W. Seeger.
\newblock Gaussian process optimization in the bandit setting: No regret and
  experimental design.
\newblock In \emph{ICML}, 2010.

\bibitem[Sun et~al.(2020{\natexlab{a}})Sun, Panda, and
  Feris]{Sun2020AdaShareLW}
Ximeng Sun, Rameswar Panda, and Rog{\'e}rio~Schmidt Feris.
\newblock Adashare: Learning what to share for efficient deep multi-task
  learning.
\newblock \emph{ArXiv}, abs/1911.12423, 2020{\natexlab{a}}.

\bibitem[Sun et~al.(2020{\natexlab{b}})Sun, Wang, Li, Feng, Tian, Wu, and
  Wang]{Sun2020ERNIE2A}
Yu~Sun, Shuohuan Wang, Yukun Li, Shikun Feng, Hao Tian, Hua Wu, and Haifeng
  Wang.
\newblock Ernie 2.0: A continual pre-training framework for language
  understanding.
\newblock \emph{ArXiv}, abs/1907.12412, 2020{\natexlab{b}}.

\bibitem[Tan et~al.(2019)Tan, Chen, Pang, Vasudevan, and Le]{Tan2019MnasNetPN}
Mingxing Tan, Bo~Chen, Ruoming Pang, Vijay Vasudevan, and Quoc~V. Le.
\newblock Mnasnet: Platform-aware neural architecture search for mobile.
\newblock \emph{2019 IEEE/CVF Conference on Computer Vision and Pattern
  Recognition (CVPR)}, pp.\  2815--2823, 2019.

\bibitem[Thoppilan et~al.(2022)Thoppilan, Freitas, Hall, Shazeer, Kulshreshtha,
  Cheng, Jin, Bos, Baker, Du, Li, Lee, Zheng, Ghafouri, Menegali, Huang,
  Krikun, Lepikhin, Qin, Chen, Xu, Chen, Roberts, Bosma, Zhou, Chang, Krivokon,
  Rusch, Pickett, Meier-Hellstern, Morris, Doshi, Santos, Duke, S{\o}raker,
  Zevenbergen, Prabhakaran, Diaz, Hutchinson, Olson, Molina, Hoffman-John, Lee,
  Aroyo, Rajakumar, Butryna, Lamm, Kuzmina, Fenton, Cohen, Bernstein, Kurzweil,
  Aguera-Arcas, Cui, Croak, Chi, and Le]{Thoppilan2022LaMDALM}
Romal Thoppilan, Daniel~De Freitas, Jamie Hall, Noam~M. Shazeer, Apoorv
  Kulshreshtha, Heng-Tze Cheng, Alicia Jin, Taylor Bos, Leslie Baker, Yu~Du,
  Yaguang Li, Hongrae Lee, Huaixiu Zheng, Amin Ghafouri, Marcelo Menegali,
  Yanping Huang, Maxim Krikun, Dmitry Lepikhin, James Qin, Dehao Chen,
  Yuanzhong Xu, Zhifeng Chen, Adam Roberts, Maarten Bosma, Yanqi Zhou,
  Chung-Ching Chang, I.~A. Krivokon, Willard~James Rusch, Marc Pickett,
  Kathleen~S. Meier-Hellstern, Meredith~Ringel Morris, Tulsee Doshi,
  Renelito~Delos Santos, Toju Duke, Johnny~Hartz S{\o}raker, Ben Zevenbergen,
  Vinodkumar Prabhakaran, Mark Diaz, Ben Hutchinson, Kristen Olson, Alejandra
  Molina, Erin Hoffman-John, Josh Lee, Lora Aroyo, Ravindran Rajakumar, Alena
  Butryna, Matthew Lamm, V.~O. Kuzmina, Joseph Fenton, Aaron Cohen, Rachel
  Bernstein, Ray Kurzweil, Blaise Aguera-Arcas, Claire Cui, Marian Croak,
  Ed~Chi, and Quoc Le.
\newblock Lamda: Language models for dialog applications.
\newblock \emph{ArXiv}, abs/2201.08239, 2022.

\bibitem[V{\'e}niat et~al.(2020)V{\'e}niat, Denoyer, and
  Ranzato]{Vniat2020EfficientCL}
Tom V{\'e}niat, Ludovic Denoyer, and Marc'Aurelio Ranzato.
\newblock Efficient continual learning with modular networks and task-driven
  priors.
\newblock \emph{ArXiv}, abs/2012.12631, 2020.

\bibitem[Wang et~al.(2019)Wang, Dai, P{\'o}czos, and
  Carbonell]{Wang2019CharacterizingAA}
Zirui Wang, Zihang Dai, Barnab{\'a}s P{\'o}czos, and Jaime~G. Carbonell.
\newblock Characterizing and avoiding negative transfer.
\newblock \emph{2019 IEEE/CVF Conference on Computer Vision and Pattern
  Recognition (CVPR)}, pp.\  11285--11294, 2019.

\bibitem[Yao et~al.(2020)Yao, Zhou, Mahdavi, Li, Socher, and
  Xiong]{Yao2020OnlineSM}
Huaxiu Yao, Yingbo Zhou, Mehrdad Mahdavi, Zhenhui~Jessie Li, Richard Socher,
  and Caiming Xiong.
\newblock Online structured meta-learning.
\newblock \emph{ArXiv}, abs/2010.11545, 2020.

\bibitem[Yoon et~al.(2018)Yoon, Yang, Lee, and Hwang]{Yoon2018LifelongLW}
Jaehong Yoon, Eunho Yang, Jeongtae Lee, and Sung~Ju Hwang.
\newblock Lifelong learning with dynamically expandable networks.
\newblock \emph{ArXiv}, abs/1708.01547, 2018.

\bibitem[Yu et~al.(2020)Yu, Kumar, Gupta, Levine, Hausman, and
  Finn]{Yu2020GradientSF}
Tianhe Yu, Saurabh Kumar, Abhishek Gupta, Sergey Levine, Karol Hausman, and
  Chelsea Finn.
\newblock Gradient surgery for multi-task learning.
\newblock \emph{ArXiv}, abs/2001.06782, 2020.

\bibitem[Zhai et~al.(2019)Zhai, Puigcerver, Kolesnikov, Ruyssen, Riquelme,
  Lucic, Djolonga, Pinto, Neumann, Dosovitskiy, Beyer, Bachem, Tschannen,
  Michalski, Bousquet, Gelly, and Houlsby]{Zhai2019TheVT}
Xiaohua Zhai, Joan Puigcerver, Alexander Kolesnikov, Pierre Ruyssen, Carlos
  Riquelme, Mario Lucic, Josip Djolonga, Andr{\'e}~Susano Pinto, Maxim Neumann,
  Alexey Dosovitskiy, Lucas Beyer, Olivier Bachem, Michael Tschannen, Marcin
  Michalski, Olivier Bousquet, Sylvain Gelly, and Neil Houlsby.
\newblock The visual task adaptation benchmark.
\newblock \emph{ArXiv}, abs/1910.04867, 2019.

\bibitem[Zhang et~al.(2011)Zhang, hui Zhan, Lin, Chen, jiao Gong, Zhong, hung
  Chung, Li, and hui Shi]{Zhang2011EvolutionaryCM}
Jun Zhang, Zhi hui Zhan, Ying Lin, Ni~Chen, Yue jiao Gong, Jinghui Zhong,
  Henry~Shu hung Chung, Yun Li, and Yu~hui Shi.
\newblock Evolutionary computation meets machine learning: A survey.
\newblock \emph{IEEE Computational Intelligence Magazine}, 6:\penalty0 68--75,
  2011.

\bibitem[Zhang et~al.(2020)Zhang, Deng, Zhang, and Wu]{Zhang2020ASO}
Wen Zhang, Lingfei Deng, Lei Zhang, and Dongrui Wu.
\newblock A survey on negative transfer.
\newblock \emph{ArXiv}, abs/2009.00909, 2020.

\bibitem[Zoph \& Le(2017)Zoph and Le]{Zoph2017NeuralAS}
Barret Zoph and Quoc~V. Le.
\newblock Neural architecture search with reinforcement learning.
\newblock \emph{ArXiv}, abs/1611.01578, 2017.

\end{thebibliography}
\bibliographystyle{iclr2023_conference}


\clearpage
\appendix


\end{document}